%% file: acl24-dialect-bias-frame.tex
\pdfoutput=1

\documentclass[11pt]{article}

\usepackage{acl}

\usepackage{times}
\usepackage{latexsym}

\usepackage[T1]{fontenc}

\usepackage[utf8]{inputenc}

\usepackage{microtype}

\usepackage{inconsolata}

\usepackage{easyReview}

\usepackage{tabularx}
\usepackage{booktabs}

\usepackage[inline]{enumitem}

\input{webis-paper-prefix}
\newcommand{\aaeweightedloss}{$\text{AAE}_{\text{wgh}}$}
\newcommand{\aaesubsampling}{$\text{AAE}_{\text{smp}}$}

\newcommand{\singletask}{\textsc{Singletask}}
\newcommand{\singletaskaae}{\textsc{Singletask}$_{\text{+AAE}}$}
\newcommand{\multitask}{\textsc{Multitask}}
\newcommand{\multitaskaae}{\textsc{Multitask}$_{\text{+AAE}}$}

\newcommand{\bspubtag}{
    \vspace*{-18.6cm}\hspace*{-0.5cm}
    {\fontsize{6}{8}\selectfont%
        \renewcommand{\arraystretch}{0.9}
        \begin{tabular}{l}
            Accepted to \\
            Findings of the Association for Computational Linguistics: ACL 2024
        \end{tabular}
}}

\begin{document}
    \input{acl24-dialect-bias-pre}

    \bspubtag
    \vspace*{17.8cm}\hspace*{0.5cm}

    \input{acl24-dialect-bias-part1}
    \input{acl24-dialect-bias-part2}
    \input{acl24-dialect-bias-part3}
    \input{acl24-dialect-bias-part4}
    \input{acl24-dialect-bias-part5}
    \input{acl24-dialect-bias-sum}

    \input{acl24-dialect-bias-limitations}
    \input{acl24-dialect-bias-ethical}
    \input{acl24-dialect-bias-acknowledgements}

    \bibliography{acl24-dialect-bias-lit}

    \appendix
    \input{acl24-dialect-bias-appendix-a}
    \input{acl24-dialect-bias-appendix-b}

    \input{acl24-dialect-bias-appendix-c}
    \input{acl24-dialect-bias-appendix-d}

\end{document}

%% file: acl24-dialect-bias-pre.tex
\title{Disentangling Dialect from Social Bias \\via Multitask Learning to Improve Fairness}

\author{
    Maximilian Spliethöver\textsuperscript{\dag}, Sai Nikhil Menon\textsuperscript{*}, and Henning Wachsmuth\textsuperscript{\dag}\\
    \textsuperscript{\dag}Leibniz University Hannover, Institute of Artificial Intelligence \\
    \textsuperscript{*}Paderborn University, Department of Computer Science \\
    \small{\tt \{m.spliethoever,h.wachsmuth\}@ai.uni-hannover.de} \\
}

\maketitle

\begin{abstract}
Dialects introduce syntactic and lexical variations in language that occur in regional or social groups. Most NLP methods are not sensitive to such variations. This may lead to unfair behavior of the methods, conveying negative bias towards dialect speakers. While previous work has studied dialect-related fairness for aspects like hate speech, other aspects of biased language, such as lewdness, remain fully unexplored. To fill this gap, we investigate performance disparities between dialects in the detection of five aspects of biased language and how to mitigate them. To alleviate bias, we present a multitask learning approach that models dialect language as an auxiliary task to incorporate syntactic and lexical variations. In our experiments with African-American English dialect, we provide empirical evidence that complementing common learning approaches with dialect modeling improves their fairness. Furthermore, the results suggest that multitask learning achieves state-of-the-art performance and helps to detect properties of biased language more reliably.
\end{abstract}

%% file: acl24-dialect-bias-part1.tex
\section{Introduction}
\label{sec:introduction}

The term \emph{social bias} is used broadly in the field of NLP. Existing works approach various facets, such as the affected social group \cite{sap2020},
the tasks for which bias is evaluated \cite{blodgett2020}, and the limited fairness of NLP systems in real-world settings, which may put specific social groups at a disadvantage while favoring others \cite{hovy2016,wu2022}.

\bsfigure{examples}{Two texts from the corpus of \citet{sap2020}, showcasing some of the five social bias aspects tackled in this paper: Neither text is \emph{lewd}, talks about some \emph{target group}, or is from an \emph{ingroup} member. Unlike (a), however, (b) is \emph{offensive} and \emph{intentional}. While (a) contains elements common in AAE, i.e., the habitual \textit{be} and dropped copula \cite{ziems2022}, (b) does not.}

A specific fairness issue arises when a bias detection model is predominantly trained and evaluated on standard language but applied to texts with \emph{dialect} \cite{jurgens2017}. Dialects appear in regional and social communities and introduce syntactic and lexical variations \cite{blodgett2016}. As such, dialects may notably diverge from the source language, posing a challenge to models trained primarily on standard language \cite{belinkov2018,ebrahimi2018,kantharuban2023}. If not explicitly accounted for, the lack of dialect understanding may subsequently lead to unfair decisions towards dialect speakers \cite{ziems2022}, originating in data and label imbalances, but also in selected dialect terms that may be considered offensive in non-dialect contexts. For example, while the use of the N-word can be acceptable when used among African-American English (AAE) speakers, its use is considered inappropriate in Standard American English (SAE) \cite{rahman2012,widawski2015,talat2018}.

Most NLP models are, however, developed without consideration for dialect patterns and may, if any, only learn them implicitly through language modeling on large corpora. For example, at the time of writing, common language models, such as GPT-3 \cite{brown2020}, LLaMA \cite{touvron2023}, Llama-2 \cite{touvron2023a}, or DeBERTaV3 \cite{he2023}, all have not explicitly been trained or evaluated for dialects. The resulting disparities between dialect and standard languages have been widely recognized \cite{blodgett2017,duarte2017,tatman2017,davidson2019,resende2024}. So far, however, few works focus on dialect-related fairness issues for social bias detection.

In this work, we investigate how to improve the fairness of social bias detection by explicitly modeling dialect language. We follow the \emph{bias} definition of \citet{sap2020}, and consider \emph{fairness} as equal classification performance between texts written in some dialect and other texts \cite{halevy2021}. Concretely, we ask:
\begin{quote}
\em How to improve fairness in automated social bias detection between dialect and non-dialect language, while maintaining detection performance?
\end{quote}

To study this question, we propose a multitask learning approach that treats the dialect detection jointly with social bias classification tasks. Our hypothesis is that, by modeling dialect as an auxiliary task, the model gains a better internal representation of dialect language patterns and bias aspects. We expect that this does not only allow it to differentiate between dialect and biased language more easily, but also benefit non-dialect texts.

For evaluation, we focus on AAE as dialect, adopting the demographic-aligned definition of \citet{blodgett2016}, and five aspects of bias, namely offensiveness, lewdness, intention, targeting a group, and being part of the target group (cf. Section~\ref{sec:experiments}). Figure~\ref{examples} illustrates how two examples relate to the dialect labels and the bias aspects.

In experiments, we compare the classification performance and fairness of our approach to baselines from related work as well as ablations that do not explicitly model dialect. We evaluate common performance and fairness metrics overall and per dialect. A perfectly fair model would show high performance without differences for dialect splits. Since, at the time of writing, no corpus for bias detection includes annotations for dialect use, we employ an automated data augmentation method.

The results of our experiments reveal performance disparities between AAE and non-AAE texts, and suggest that modeling multiple bias aspects helps to detect biased language more reliably. The proposed multitask learning approach improves over the best baseline and over single-task learning for four out of five bias aspects. Moreover, the dialect auxiliary task improves fairness for texts with dialect language and also benefits non-dialect texts. Learning five bias aspects and dialect detection simultaneously shows the most stable fairness and performance improvements across tasks.

To summarize, our main contributions are:
\begin{enumerate}
    \item
    A multitask learning approach to jointly learn dialect and social bias detection.
    \item
    Evidence that simultaneously modeling multiple bias aspects and dialect language improves the classification performance and fairness for (AAE) dialect speakers.%
    \footnote{Code at: \url{https://github.com/webis-de/ACL-24}}
\end{enumerate}

%% file: acl24-dialect-bias-part2.tex
\section{Related Work}
\label{sec:relatedwork}

Social bias can be defined as stereotypical thinking or prejudices against social groups \cite{fiske1998}. In NLP, it can manifest in hidden representations \cite{spliethover2020} or unfair predictions \cite{angwin2016}. Identifying social bias in data is an important step towards debiasing NLP models, since models adopt and amplify pre-existing biases which can have harmful effects \cite{zhao2017,shwartz2020}.

In related work, %
\citet{wald2023} analyze bias in fine-tuned large language models (LLM) as a proxy for bias in data. Focusing on single texts, \citet{sap2020} introduce the Social Bias Inference Corpus (SBIC) and train a model to predict multiple bias aspects. \citet{prabhumoye2022} apply few-shot learning to instruction-tuned LLMs on the same data. We build on the work of \citet{sap2020}, but extend it by considering dialects.

Relevant to dialects, another perspective to social bias is fairness regarding performance across social groups \cite{tolan2018}. For example, \citet{tatman2017} find that video captioning systems perform worse for dialect speakers and women. In NLP, \citet{blodgett2016} highlight disparities between dialect and standard language, as well as social groups in language identification. \citet{resende2024} find negative biases in hate speech detection towards AAE texts due to underrepresentation in datasets. However, no work so far has considered fairness for the interplay of dialects and social bias. \citet{ziems2022} find, similar to \citet{joshi2024}, that models perform worse on AAE compared to SAE in natural language understanding tasks.

Many existing fairness evaluations target toxicity detection. \citet{mozafari2020} observe that a fine-tuned model labels AAE texts more often as hate speech than SAE texts and propose a mitigation strategy. Similarly, \citet{halevy2021} aim to mitigate this bias by introducing a dialect detection and dialect-specific toxicity classifiers. Their ensemble reconsiders positive toxicity predictions for AAE texts. \citet{badjatiya2019} show that hate speech detection can be improved by removing bias-sensitive words commonly used in dialects, and \citet{xia2020} propose an adversarial model to prevent false classifications for AAE text.

We aim to identify multiple aspects of biased language rather than just toxicity, and intend to incorporate dialect language into the model using multitask learning instead of separate models.
Closest to our approach, \citet{talat2018} try to overcome socio-demographic differences in hate speech annotations that arise from diversity in contexts and definitions. They train a multitask learning model on texts from various domains, annotated with separate definitions for hate speech. In experiments, their model outperforms existing approaches and generalizes better to unseen data. Unlike us, however, they do not account for dialect language, nor evaluate for respective social groups. Moreover, we use multitask learning to explicitly incorporate socio-demographic knowledge, namely dialect language patterns, into a unified model.

For data availability reasons, we automatically augment the SBIC. Modern language models have been tested with respect to their capabilities to augment training data. For example, \citet{faggioli2023} evaluate the utility of LLMs as annotators for supervised learning and find that, for relevance judgment tasks, automatically-generated annotations show promising results. \citet{zhang2023} extend the idea by introducing active learning. In contrast, we fine-tune an encoder model on a separate AAE dialect dataset and use it to annotate the bias aspect dataset for dialect language usage.

In terms of dialect data, \citet{ziems2022}, propose a rule-based approach to create a new benchmark derived from GLUE \cite{wang2018}. \citet{blodgett2016} introduce the TwitterAAE dataset containing around 60 million tweets with semi-supervised annotations. While many corpora focus on AAE vs. SAE \cite{groenwold2020}, the TwitterAAE annotations make no further assumption about the non-AAE text. While a notable portion may be SAE, some might use other dialects. Since our work focuses on AAE vs. non-AAE and on texts from the internet, we employ the TwitterAAE corpus to train a dialect classifier.

%% file: acl24-dialect-bias-part3.tex
\section{Methodology}
\label{sec:method}

This work focuses on making model classifications fairer for dialect texts. In the following, we present our approach to improve fairness by integrating dialect with bias aspect detection in a multitask learning architecture. As dialect annotations for bias detection data are unavailable so far, we describe how we augment existing data automatically.

\subsection{Joint Modeling of Social Bias and Dialect}
Research has shown that a primary task's performance can improve through multitask learning, in which a trained model can transfer knowledge between primary and auxiliary tasks \cite{caruana1998}.

In this work, we hypothesize that learning to detect dialect language as an auxiliary task improves the performance of a primary task for texts written in the given dialect. The addition of dialect aids the model in distinguishing simple dialect use from actual bias markers. For example, the text ``\emph{We was at some random-ass bar}'' \cite{ziems2022} might not be lewd in a context where the dialect is commonly used, such as conversations between AAE speakers. However, similar use of the word \emph{ass} in non-AAE contexts might be perceived as lewd or obscene \cite{ziems2022}.

Furthermore, we expect that multitask learning does not only improve fairness, but also the reliability of identifying bias aspects. Given that bias aspects such as offensiveness, intentionality, and targeting a group are not fully independent from each other, multitask learning may leverage interdependencies to make more accurate predictions.

To operationalize our hypotheses, we propose a weight-sharing joint learning architecture \cite{collobert2008} that uses a shared encoder and separate classification heads for each task using a standard cross-entropy loss \cite{jurafsky2021}, computed for each sample separately:
\begin{eqnarray}
L_(\hat{\mathbf{y}},\mathbf{y}) & :=	& -\sum_{i=1}^n y_i \log \hat{y}_i,
\end{eqnarray}
where $n$ is the number of samples, $y_i$ the true label, and $\hat{y}_i$ the softmax output at position $i$. Here, the used labels alternate between the different dialect and bias aspect tasks in a round-robin manner.

Figure~\ref{approach-abstract} illustrates our joint multitask learning architecture with $k$ task-specific classification heads. We further add a classification head for the auxiliary task of learning to detect a specific dialect. At training time, each head is conditioned on one task (i.e., bias aspect and dialect), whereas the shared encoder is fine-tuned for all $k+1$ tasks. In our specific case, the loss for the encoder model is calculated by alternating round-robin between tasks and individually being backpropagated to the encoder. For inference, only the classification head for the primary task is used.%
\footnote{While joint learning requires re-training for new tasks or dialects, continual learning \cite{phang2019,scialom2022} faces similar issues, and adapter fusion \cite{pfeiffer2021} performed notably worse in preliminary tests.}

\subsection{Data Augmentation with Dialect Labels}
Previous work has shown that parallel data benefits multitask learning, as correlations between multiple labels are easier to identify, positively affecting all learned tasks \cite{pfeiffer2020}. As noted in Section~\ref{sec:relatedwork}, however, no bias corpus with dialect labels exist. Furthermore, relying on a separate corpus for the auxiliary task \cite{collobert2008,talat2018} may easily cause domain transfer problems and introduce noise.%
\footnote{Preliminary tests confirmed this assumption.}

Therefore, we employ a data augmentation method. To this end, we train a dialect classifier on a separate corpus and then add dialect labels to the main corpus, as detailed in Section~\ref{sec:experiments}. Augmenting an existing corpus has two main advantages:%
\footnote{Data augmentation also makes the experiments more controlled and less dependent on the content of the dialect corpus.}
\begin{enumerate}
    \setlength{\itemsep}{0pt}
    \item
    It enables multitask learning approaches to transfer knowledge between the primary and auxiliary tasks more efficiently.
    \item
    It is more generally applicable to other dialects, since dialect-specific classifiers can be developed independent from the approach.
\end{enumerate}

\bsfigure{approach-abstract}{Our joint learning architecture: \emph{Dialect classification} is added as an additional head to the classification of the bias aspects. During training, all classification heads are trained round-robin in alternating manner. For inference, only the classification head of the primary task is used, here the \emph{Bias aspect $\#k$} head.}

%% file: acl24-dialect-bias-part4.tex
\section{Experiments}
\label{sec:experiments}

This section evaluates the effectiveness of multitask learning in improving fairness for dialects. We focus on African-American English (AAE) as a dialect and five social bias classification tasks. Below, we describe the experimental setups of our dialect data augmentation and our social bias detection approach. Using the augmented data, we test whether multitask learning improves fairness for dialect speakers if the dialect is modeled explicitly.

\subsection{Data}
\label{sec:data}

We use the following two corpora for the auxiliary task of dialect classification and for the primary task of social bias classification, respectively.

\paragraph{TwitterAAE Corpus}
We train and evaluate AAE classification on the TwitterAAE corpus \cite{blodgett2016}. The corpus contains about 59 million tweets from 2013, of which around 1 million are labeled as AAE dialect (dataset statistics are reported in Appendix~\ref{sec:appendix-data-details}). The annotation was done semi-supervised, partially based on geolocation and user demographics.
Since \citet{blodgett2016} do not provide data splits, we randomly seperate 80\% as training and validation set, and 20\% as test set, in a stratified way to preserve dialect label ratios (using a seed for reproducability).

\paragraph{Social Bias Inference Corpus}
The Social Bias Inference Corpus (SBIC) \cite{sap2020} consists of about 45k English posts from online communities. Humans annotated each post for different aspects of biased language, which include the following five classification labels:%
\footnote{Additionally, the corpus includes two free-text annotations describing the specific social group being targeted and the implied statement of a text. Since we focus on classification fairness, we disregard the free-text labels in this work.}
\begin{enumerate}
\setlength{\itemsep}{-1.1pt}
    \item \emph{Offensiveness.} Whether or not a text is rude, disrespectful or shows toxicity
    \item \emph{Intent.} If a text is offensive, whether this offensiveness was intentional or not
    \item \emph{Lewdness.} Whether or not a text contains obscene or sexual references
    \item \emph{Target Group.} Whether or not a text is directed towards a specific social group
    \item \emph{Ingroup.} Whether or notthe author of the text is part of the targeted social group%
\end{enumerate}

We use the aggregated version of SBIC, for deduplicated texts and preprocessed labels (dataset statistics in Appendix~\ref{sec:appendix-data-details}). For preprocessing, we lowercase texts, and remove mentions, retweet markers, URLs, and non-English characters. Following \citet{sap2020}, we binarize all labels.

\subsection{African-American English Classification}

To augment the SBIC with AAE dialect annotations, we develop a classifier to identify AAE texts. As mentioned in Section~\ref{sec:relatedwork}, we explicitly refrain from distinguishing AAE and SAE, as our goal is to seperate AAE from non-AAE texts.

\paragraph{Approach}

We fine-tune DeBERTa-v3-base \cite{he2023} with a classification head on the TwitterAAE corpus. While bigger models exist, BERT-based text encoders still show state-of-the-art performance in various downstream tasks \cite{he2023} and remain competitive for text-only classification tasks \cite{chen2023}. Furthermore, the DeBERTa-v3-large variant did not show a notable increase in performance in preliminary tests.

Due to the strong imbalance in the TwitterAAE corpus, we evaluate two training methods:%
\footnote{In early tests, just fine-tuning led to a majority classifier.}
\begin{itemize}
\setlength{\itemsep}{0pt}
\item
\emph{Subsampling (\aaesubsampling{}).} We randomly sample non-AAE texts in the training data (using a seed for reproducability) to match the number of AAE texts and create a balanced dataset. This method aims to equalize the importance of AAE and non-AAE labels.
\item
\emph{Loss weighting (\aaeweightedloss{}).} Subsampling removes a potentially large number of training instances (nearly 57 million in this case). Instead, this method weighs the loss of each label, relative to the label distribution. For the given data, wrongly (or correctly) classifying AAE texts has, therefore, a higher impact on the model weights during backpropagation.
\end{itemize}

\paragraph{Baselines}

We compare our approach to the \emph{TwitterAAE} dialect classifier presented by \citet{blodgett2016}. To verify that the models show a learning effect, we also report \emph{majority} and \emph{random} classifiers. Since we aim to reliably find AAE texts in particular, we emphasize the recall for this class.

\paragraph{Measures}

For all models, we report per-class and macro-averaged precision, recall, and F$_1$-scores.

\subsection{Social Bias Detection}

Now, we detail our proposed multitask learning approach to social bias detection, ablations to further investigate fairness, and the baselines we compare to. All proposed models are based on the DeBERTa-v3-base encoder \cite{he2023}.%
\footnote{For results with RoBERTa-base as encoder model, see Appendix~\ref{sec:appendix-classification-results}.}

\paragraph{Approach}
We train a joint-weight multitask learning model (\multitaskaae{}) that consists of a shared encoder and separate classification heads for the five bias aspects. Moreover, we add a classification head to detect AAE dialect. We, therefore, train a model with six classification heads in total. %
As detailed above, the additional dialect detection task should help the model to better differentiate between dialects and biased language properties. Theoretically, this could be extended to further dialects, but we chose to restrict this study to AAE as an example.

\paragraph{Ablations}
To examine the effect of the auxiliary tasks, we evaluate two ablations of the approach. First, we train the multitask learning model on the five classification tasks \emph{without}  AAE dialect modeling (\multitask{}). This allows us to analyze the auxiliary task's influence on the multitask learning model. Second, we train single-task models \emph{with} (\singletaskaae{}) and \emph{without} (\singletask{}) AAE detection to better understand its influence on each task. To do so, we use a similar joint-weights learning setup as above.%
\footnote{While the \singletaskaae{} is, in fact, also a multitask learning model, we refer to it as single-task model with AAE for clarity and better differentiation to the proposed approach.}

\paragraph{Baselines}
We compare our approaches and ablations to two baselines from related works for overall performance. First, we use the results of the best approach of \citet{sap2020}. The authors employ and fine-tune a \emph{GPT-2} model, formulating the problem as an auto-regressive generation task. Second, we use scores of the overall best approach reported by \citet{prabhumoye2022}. Like \citet{sap2020}, the authors formulate the task as a generation task, but do so in a Q\&A format. Instead of fine-tuning, however, \citet{prabhumoye2022} employ a \emph{few-shot} learning setup, providing the model with in-context examples during inference. Since neither the code, the model, nor the predictions per text are available, we compare to the scores reported in the respective papers.

\paragraph{Measures}
Following \citet{sap2020} and \citet{prabhumoye2022}, we report the positive-class F$_1$-score for each task to assess classification performance.%
\footnote{For completeness, we also report negative class and macro-averaged results in Appendix~\ref{sec:appendix-classification-results}.}
To study potential disparities and improvements, we further evaluate the classification performance per dialect (i.e., AAE and non-AAE). This allows us to test our hypothesis that classifiers work better for non-AAE texts than AAE texts and if the proposed approach improves upon this. Lastly, we also consider two common fairness metrics \cite{garg2020}: \textit{Predictive parity} describes the delta between both groups' precision scores (in this context referred to as the positive predictive value). Predictive parity is said to be satisfied if it is $0$. \textit{Equalized odds}, on the other hand, describes fairness based on recall (in this context referred to as true positive rate) and the false positive rate. It is said to be satisfied if deltas between the dialect groups are $0$.

%% file: acl24-dialect-bias-part5.tex
\section{Results and Discussion}
\label{sec:results}

We first discuss the results of dialect classification, before we look at its interplay with social bias.

\input{table-results-aae-classification-subsampled}

\subsection{African-American English Classification}

The results of the African-American English (AAE) dialect classification are reported in Table~\ref{tab:results-aae-classification-subsampled}. Due to the heavy imbalance of the test dataset (only 2\% are labeled as AAE dialect), we report metric scores on a randomly subsampled test set that balances AAE and non-AAE texts (using a seed for reproducability), with around 230k samples per class. For completeness, results on the full test set are reported in Appendix~\ref{sec:appendix-classification-results}.

Both our approaches, \aaeweightedloss{} and \aaesubsampling{}, outperform the previous state-of-the-art approach TwitterAAE in nearly all evaluations significantly. Especially, the gains on positive class precision (.80 vs.\ .73 for \aaeweightedloss{}) and recall (.74 vs.\ .64 for \aaeweightedloss{}) are noteworthy, as they allow us to identify more actual AAE texts more reliably in our main analysis. While scores increased less substantial over the baseline in negative class recall (i.e., identifying more non-AAE texts correctly, with .81 vs.\ .77 for \aaeweightedloss{}), increases in negative class precision are similarly noteworthy (.76 vs.\ .68 for \aaeweightedloss{}).

Overall, \aaeweightedloss{} not only performs better than TwitterAAE in all metrics, but also improves over \aaesubsampling{}, except for positive class recall and negative class precision. While the recall of the positive class is important in this task, \aaesubsampling{} would likely introduce more noise through false predictions, as indicated by its lower recall for the negative class, which also does not improve over the baseline. Therefore, we use \aaeweightedloss{} to augment the SBIC data with AAE dialect annotations.

\input{table-results-bias-classification}

\input{table-results-per-dialect-bias-classification}

\subsection{Social Bias Detection}

Table~\ref{tab:results-bias-classification} presents the results of predicting the five bias aspects. Following \citet{sap2020} and \citet{prabhumoye2022}, we report the F$_1$-score of the positive class for each aspect (for negative and macro F$_1$, see Appendix~\ref{sec:appendix-classification-results}).

Fine-tuning on single labels seems to work notably better than using a generative approach: \singletask{} outperforms the two baselines (GPT-2 and few-shot learning) on three bias aspects significantly. We observe a strong F$_1$-score gain of 9.6 points over the best baseline on the target group aspect (.833 vs.\ .737).
A better dialect language understanding (\singletaskaae{}) further improves performance on lewdness from .744 to .755, while it seems to not influence the performance on offensiveness and intent. These results indicate that AAE dialect texts are most impacted by wrong predictions on the lewdness aspect when finetuning on a single task. In a qualitative analysis, we find that particularly for lewdness, better knowledge of AAE dialect patterns is helpful (cf. Section~\ref{sec:results-qualitative-analysis}). Finally, all approaches that do not explicitly model AAE dialect predict the majority label for ingroup.\,\,\,

Both multitask approaches further improve upon \singletask{}, showing performance increases in most aspects except target group. The biggest gains are achieved for the offensiveness and ingroup aspects. For ingroup, seemingly the most challenging aspect, \multitask{} and \multitaskaae{} are among the only three evaluated models that show a learning effect, with a significantly improved F$_1$-score of .235 and .227 respectively. This may be due to a better ability to detect non-offensive contexts containing terms usually used offensively, and an increased awareness of impossible label combinations, such as predicting that an author is part of the target group, while also predicting that no group was targeted (cf. Section~\ref{sec:results-qualitative-analysis}). These results indicate that multitask learning helps to detect biased aspects of language. Moreover, especially for more complex and implicit signals, such as ingroup, considering several aspects can help.

Modeling dialects seems to improve results most when finetuning on a single task. This is most visible for lewdness and ingroup, where the scores of \singletaskaae{} increase by .011 and .108 over \singletask{}, respectively. A reason may be that jointly modeling the task and dialect disentangles dialect language from bias aspects and explicit word use, resulting in a model with better conceptual representations of the respective aspect (cf. Section~\ref{sec:results-qualitative-analysis}). A model that considers only the aspect might not be complex enough to correctly interpret subtle changes in language introduced by dialects and thus maybe more prone misclassifications. Interestingly, \multitaskaae{} does not benefit in the same of from modeling the dialect in addition to multiple aspects.

In conclusion, we find that, while supervised classification shows a notable performance increase over generative approaches in detecting bias aspects, considering multiple aspects of biased language jointly, clearly improves the reliability of the predictions further. Modeling dialect in addition to bias aspects has the most impact on models that consider only a single aspects otherwise.

\subsection{Fairness in Social Bias Detection}

Table~\ref{tab:results-per-dialect-bias-classification} shows the results of bias aspect detection for AAE and non-AAE texts. As hypothesized in Section~\ref{sec:introduction}, the simple supervised fine-tuning of \singletask{} performs indeed better for non-AAE texts, showing a difference of up to 10 points (.863 vs.\ .755 for target group). Such disparities could severely impact fairness if deployed in real-world applications. For example, if a system automatically flags offensive posts for removal, posts by dialect speakers would be falsely removed more often due to the decreased ability of the classifier to model dialect language.

The disparities between AAE and non-AAE performance further suggest that the \singletask{} approach still partially relies on word usage rather than meaning to identify certain bias aspects, showing limited awareness of the AAE dialect (see Section~\ref{sec:results-qualitative-analysis}). This interpretation is further supported by the fact that the \singletaskaae{} ablation improves the performance on AAE texts for most aspects, and especially offensiveness and intent. While the gains come with decreases on non-AAE texts for selected aspects, dialect modeling still reduces the performance gap between AAE and non-AAE most of the time. It may thus be seen as a worthy trade-off, depending on the application.

Interestingly, however, \singletaskaae{} shows the opposite effect for two aspects: For lewdness, the performance on AAE texts drops from .842 to .836, but increases for non-AAE texts from .861 to .869. Similarly, an increase for non-AAE texts is visible for the target group and ingroup aspects. These results indicate that awareness of dialect language helps improves results for texts written with dialect, but also for those without.

The gains of multitask learning over the single-task approaches on AAE texts are more consistent. \multitask{} and \multitaskaae{} reach the most notable gains on AAE texts for offensiveness and ingroup, improving upon \singletask{} (.787) to .816 and .806, respectively. In terms of fairness, AAE modeling shows a similar pattern for multitask and single-task learning. \multitaskaae{} improves the performance on AAE texts over \multitask{} and \singletask{} on selected aspects. Unlike for \singletaskaae{}, the gain in performance is only visible for AAE, while often slightly decreases for non-AAE. On lewdness, for example, the score increases significantly from .840 (\multitask{}) to .846 (\multitaskaae{}) for AAE texts, but decreases from .870 (\multitask{}) to .865 (\multitaskaae{}) for non-AAE texts. Similarly, the scores improve from .630 to .639 for ingroup on AAE texts. This supports our hypothesis that modeling dialects can improve internal concept representations of bias aspects (cf. Section~\ref{sec:results-qualitative-analysis}), potentially being more important for singletask than for multitask approaches.

\input{table-results-fairness-offensive}

\paragraph{Equalized Odds and Predictive Parity}

\multitaskaae{} has the biggest impact for {equalized odds} and {predictive parity}. Table~\ref{tab:results-fairness-offensive} shows the results exemplarily for \textit{offensiveness}.%
\footnote{Other aspects show similar patterns (see Appendix~\ref{sec:appendix-classification-results}).}
For \singletaskaae{}, the dialect modeling improves classification performance for AAE dialect most of the time, and also lowers the performance gap between AAE and non-AAE texts compared to \singletask{}. While not all gains are substantial, they are, again, consistent across metrics and bias aspects. We hence see these results as further evidence for a positive impact of dialect modeling, especially in singletask learning architectures.

\medskip
Overall, both multitask learning and dialect modeling seem to improve the performance of texts written in a given dialect, as we displayed for AAE dialect in this work. The evaluation also suggests that the proposed models consistently make fairer predictions. While we focus on AAE dialect in this study, the dialect classification and bias detection component can be adapted to other dialects. Since the two components are independent, there are no constraints on the chosen dialect classification approach. We therefore expect that modeling dialects may also improve language understanding and performance for dialects other than AAE and encourage future work to consider this direction.

\subsection{Qualitative Analysis}
\label{sec:results-qualitative-analysis}

To further investigate potential improvements of modeling dialect as an auxiliary task, we conducted a qualitative analysis. Here, we summarize the main results only. A more exhaustive version of the analysis, including specific examples from the data, can be found in Appendix~\ref{sec:appendix-qualitative-analysis}.

\paragraph{Latent Bias Concepts}

Generally, modeling AAE dialect seems to benefit offensiveness and lewdness classification by improving the internal concept representations of the respective aspects. These improved representations seem to help the model abstract from word use towards relying more on contextual information. The better abstraction further seems to improve the interpretation of the whole context to decide whether a text is written by an ingroup member, as also demonstrated in Table~\ref{tab:results-per-dialect-bias-classification}.

\paragraph{Implicit Label Dependencies}

Due to the conceptual dependency of some labels in the SBIC corpus, certain label-value combinations are nonsensical and should not be predicted. For example, while a text can be intentionally offensive (thus being labeled as \emph{offensive} and \emph{intentional}, by definition of the aspects, it is impossible for a text to be intentional but not offensive. To model and identify this connection, however, it is necessary to consider both aspects simultaneously. While the \singletask{} model predicts these impossible combinations for \emph{offensive} and \emph{intentional} only 59 times, both multitask learning variants, \multitask{} and \multitaskaae{}, eliminate the issue and never predict such wrong combinations. A similar effect can be observed for the \emph{target group} and \emph{ingroup} labels. These insights highlight the benefit of considering multiple bias aspects together. Future work might further investigate this effect.

%% file: table-results-aae-classification-subsampled.tex
\begin{table}[t]
    \small
    \centering
    \setlength{\tabcolsep}{2pt}

    \begin{tabular}{lrrrrrrrrr}
        \toprule
        & \multicolumn{3}{c}{\textbf{Precision $\uparrow$}} & \multicolumn{3}{c}{\textbf{Recall $\uparrow$}} & \multicolumn{3}{c}{\textbf{F$_1$ $\uparrow$}}  \\
        \cmidrule(l@{3pt}r@{2pt}){2-4}\cmidrule(l@{3pt}r@{2pt}){5-7}\cmidrule(l@{3pt}){8-10}
        \textbf{Model} & \textbf{Pos} & \textbf{Neg} & \textbf{Mac} & \textbf{Pos} & \textbf{Neg} & \textbf{Mac} & \textbf{Pos} & \textbf{Neg} & \textbf{Mac} \\
        \midrule

        Majority & .00 & .50 & .25 & .00 & \bf 1.0 & .50 & .00 & .67 & .33 \\

        Random & .50 & .50 & .50 & .50 & .50 & .50 & .50 & .50 & .50 \\
        [.5em]

        Twi.AAE & .73 & .68 & .71 & .64 & .77 & .70 & .69 & .72 & .70 \\
        [.5em]

        \aaeweightedloss & \ddag\bf .80 & \ddag.76 & \bf \ddag.78 & \ddag.74 & \ddag.81 & \ddag\bf .78 & \ddag\bf .77 & \ddag\bf .78 & \ddag\bf .78 \\

        \aaesubsampling & \ddag.77 & \ddag\bf .78 & \ddag.77 & \ddag\bf .78  & .76 & \ddag.77 & \bf \ddag.77 & \ddag.77 & \ddag.77 \\

        \bottomrule
    \end{tabular}

    \caption{African-American English dialect classification results: postive class (\textit{Pos}), negative class (\textit{Neg}) and macro-averaged (\textit{Mac}). While \aaesubsampling{} seems better in finding dialect texts (higher recall for \textit{Pos}), \aaeweightedloss{} performs better overall. Gains of both approaches over the TwitterAAE baseline are significant (\ddag{} for $p<.01$).}
    \label{tab:results-aae-classification-subsampled}
\end{table}

%% file: table-results-bias-classification.tex
\begin{table}[t]
    \small
    \renewcommand{\arraystretch}{1}
    \centering
    \setlength{\tabcolsep}{3.4pt}
    \begin{tabular}{lrrrrr}
        \toprule
        \textbf{Model} & \textbf{Offens.} & \textbf{Intent} & \textbf{Lewdn.} & \textbf{Target} & \textbf{Ingroup} \\
        \midrule
        Majority & .732 & .694 & .000 & .000 & .000 \\
        Random & .529 & .504 & .165 & .456 & .038 \\
        [.5em]
        GPT-2 & .788 & .786 & \textbf{.807} & .699 & .000 \\
        Few-shot & .822 & .798 & .411 & .737 & -- \\
        [.5em]
        \textsc{Stl} & \ddag.875 & \ddag.861 & .744 & \ddag.832 & .000 \\
        \textsc{Stl}$_{\text{+AAE}}$ & \ddag.875 & \ddag.861 & .755 & \ddag\textbf{.833} & .108 \\
        [.5em]
        \textsc{Mtl} & **\ddag\textbf{.882} & **\ddag\textbf{.864} & **.757 & \ddag.832 & **\ddag\textbf{.235} \\
        \textsc{Mtl}$_{\text{+AAE}}$ & *\ddag.879 & \ddag\textbf{.864} & .751 & \ddag.831 & **\ddag.227 \\
        \bottomrule
    \end{tabular}
    \caption{Bias classification results (positive-class F$_1$, averaged over five random seeds) for each aspect: offensiveness, intent, lewdness, target group, and ingroup. The additional dialect modeling in \multitaskaae{} (\textsc{Mtl}$_{\text{+AAE}}$) improves over single-task approaches and baselines in most cases. Most gains over the strongest baseline per bias aspect are significant (\ddag{} for $p < .01$). Significant gains of multitask approaches over single-task variants are marked by * ($p< .05$) or ** ($p< .01$).}
    \label{tab:results-bias-classification}
\end{table}

%% file: table-results-per-dialect-bias-classification.tex
\begin{table*}[t]
    \small
    \renewcommand{\arraystretch}{1}
    \centering
    \setlength{\tabcolsep}{6.8pt}

    \begin{tabular}{lrrrrrrrrrr}
        \toprule
        & \multicolumn{2}{c}{\textbf{Offensiveness}} & \multicolumn{2}{c}{\textbf{Intent}} & \multicolumn{2}{c}{\textbf{Lewdness}} & \multicolumn{2}{c}{\textbf{Target Group}} & \multicolumn{2}{c}{\textbf{Ingroup}}  \\
        \cmidrule(l@{6pt}r@{3pt}){2-3}\cmidrule(l@{6pt}r@{3pt}){4-5}\cmidrule(l@{6pt}r@{3pt}){6-7}\cmidrule(l@{6pt}r@{3pt}){8-9}\cmidrule(l@{6pt}){10-11}
        \textbf{Model} & \textbf{$\lnot$AAE} & \textbf{AAE} & \textbf{$\lnot$AAE} & \textbf{AAE} & \textbf{$\lnot$AAE} & \textbf{AAE} & \textbf{$\lnot$AAE} & \textbf{AAE} & \textbf{$\lnot$AAE} & \textbf{AAE} \\
        \midrule
        Majority & .361 & .397 & .344 & .368 & .476 & .467 & .372 & .360 & .497 & .482 \\
        Random & .495 & .500 & .492 & .487 & .396 & .444 & .499 & .499 & .342 & .370 \\
        [.5em]
        \singletask & .854 & .787 & .854 & .786 & .861 & .842 & .863 & .755 & .497 & .482 \\
        \singletaskaae & .851 & \ddag{}.808 & .853 & \bf .790 & \dag{}.869 & .836 & \bf .867 & \bf .760 & .530 & .550 \\
        [.5em]
        \multitask & **\bf .860 & **\bf .816 & *\bf .856 & .784 & **\bf .870 & .840 & \bf .867 & .756 & ** \bf .569 & **.630 \\
        \multitaskaae & *.856 & .806 & .855 & .783 & .865 & *\ddag{}\bf .846 & \bf .867 & .749 & .553 & \bf .639 \\
        \bottomrule
    \end{tabular}

    \caption{Bias classification results (macro F$_1$, averaged over five random seeds) for texts with and without AAE dialect (see Appendix~\ref{sec:appendix-classification-results} for precision, recall, and F$_1$-scores per class). While multitask learning has a notable impact, the AAE modeling especially improves the performance of singletask models. Significant gains are marked for multitask models over single-task variants (* $p<.05$, ** $p<.01$) and for AAE models over those without AAE modeling (\dag{} $p<.05$, \ddag{} $p<.01$).}
    \label{tab:results-per-dialect-bias-classification}
\end{table*}

%% file: table-results-fairness-offensive.tex
\begin{table}[t]
    \small
    \centering
    \setlength{\tabcolsep}{2.85pt}

    \begin{tabular}{lrrrrrr}
        \toprule

        \bf Offens. & \multicolumn{2}{c}{\textbf{TPR $\uparrow$}} & \multicolumn{2}{c}{\textbf{FPR $\downarrow$}} & \multicolumn{2}{c}{\textbf{PPV $\uparrow$}} \\

        \cmidrule(l@{3pt}r@{2pt}){2-3}\cmidrule(l@{3pt}r@{2pt}){4-5}\cmidrule(l@{3pt}){6-7}
        \textbf{Model} & \textbf{$\lnot$AAE} & \textbf{AAE} & \textbf{$\lnot$AAE} & \textbf{AAE} & \textbf{$\lnot$AAE} & \textbf{AAE} \\

        \midrule

        \textsc{Stl} & .893 & .918 & .190 & .372 & .860 & .826 \\

        \textsc{Stl}$_{\text{+AAE}}$ & .891 & \ddag{}\bf .938 & .193 & \ddag{}.354 & .857 & \ddag{}.836 \\
        [.5em]

        \textsc{Mtl} & \bf .896 & *.934 & \bf .181 & **\bf .331 & *\bf .866 & **\bf .845 \\

        \textsc{Mtl}$_{\text{+AAE}}$ & \bf .896 & .935 & .189 & .353 & .861 & .836 \\

        \bottomrule
    \end{tabular}

    \caption{Fairness results per approach in terms of true positive rate (\emph{TPR}), false positive rate (\emph{FPR}), and positive predictive value (\emph{PPV}) for the offensiveness aspect (averaged over five random seeds; other aspects in Appendix~\ref{sec:appendix-classification-results}). Multitask learning improves results for AAE and reduces some differences to non-AAE (\emph{$\lnot$AAE}). \multitask{} is best in most regards. Significant improvements are marked for multitask models over single-task variants (* $p<.05$, ** $p<.01$) and for AAE over non-AAE models (\dag{} $p<.05$, \ddag{} $p<.01$).}
    \label{tab:results-fairness-offensive}
\end{table}

%% file: acl24-dialect-bias-sum.tex
\section{Conclusion}
\label{sec:conclusion}

In this work, we have studied the fairness of social bias detection with respect to dialects, presenting a multitask learning approach to model dialect as an auxiliary task. The approach aims to mitigate disparities in the bias detection performance across dialects. For data availability reasons, our experiments have focused on African-American English (AAE) texts and non-AAE texts.
We have obtained state-of-the-art performance in predicting five different aspects of social bias. Moreover, modeling AAE dialect as an auxiliary task narrows selected performance disparities learning setup, thus making results fairer for AAE dialect texts.

This work, therefore, provides empirical evidence that explicitly encoding dialect language patterns into models can have a positive impact on fairness for dialect speakers. Especially when coupled with a reliable dialect classification model, we expect a notable effect. In future work, we aim to investigate how to model dialect language to improve our data augmentation methods, i.e., using as automatic dialect translation methods \cite{ziems2022} and counterfactual data generation \cite{zmigrod:2019,stahl:2022}. Lastly, collecting dialect data from more diverse sources should help to scale our approach to further text genres.

We hope this paper contributes towards fairer bias classification, and encourage others to consider dialects more broadly for NLP applications.

%% file: acl24-dialect-bias-limitations.tex
\section*{Limitations}
\label{sec:limitations}

To draw conclusions about dialect and standard language, it is essential to not only consider a single dialect. We, therefore, aimed to be careful in our work to point out that our results are limited to AAE vs. non-AAE language. A potential improvement could be to introduce a three-class classifier that can classify ``AAE,'' ``Standard,'' and ``Other'' language. More preferably, though, one could incorporate more dialects than just AAE. However, given that no part other than the dialect classifier is specific to AAE, we think that our experiments could be repeated for other dialects.

A second limitation concerns the AAE dialect annotations themselves. Since SBIC does not have such annotations, we have labeled them automatically using our classifier. Due to the fact that the fairness evaluation relies on the quality of the AAE dialect classifier, our results might be less conclusive than if humans had annotated the data. The same applies to all approaches incorporating the AAE dialect classifier for their predictions. While the AAE dialect classifier is far from perfect, we still consider it rather reliable based on our evaluation and think it provides a reasonable basis for our analysis, even if it cannot be conclusive.

Lastly, our results might be limited by the fact that not only our final evaluation dataset, SBIC, but also the AAE dialect dataset considers texts exclusively from online platforms (Twitter and Reddit) and in the English language. As mentioned earlier, dialects appear and vary in regional and social communities. Our evaluations therefore investigate only a sub-group of AAE speakers. Additionally, texts from online platforms usually show language patterns very different from other text forms, such as books or news articles \cite{nguyen2020}. However, we consider this to be only a minor limitation since texts from online platforms are often used as resources for many kinds of NLP models. Reliably and fairly identifying social bias in such data is thus important and necessary. One just has to be aware that the approaches in this work might not apply to other forms of text in the same way.

%% file: acl24-dialect-bias-ethical.tex
\section*{Ethical Considerations}

With our work, we try to improve selected ethical aspects of NLP applications. Namely, we consider the case of social bias detection with a specific focus on fairness for texts written in dialect language. If the developed approaches work as intended, they should make overall predictions on social bias detection fairer for members of dialect-speaking social groups. In our specific case, those are members of the African-American community that choose to write in the AAE dialect. However, since the approaches and evaluation in this work focus on AAE dialect, they disregard other dialects and also Standard American English to a certain degree. Also, since we base our AAE dialect classifier on the dataset of \cite{blodgett2016}, we limit the classifier ability to detect dialect language patterns present in the data. We acknowledge that other variants and of AAE dialect exist for regional and social communities that were not included in the data, i.e., because they do not make (extensive) use of Twitter. In both cases, the classification performance might be implacted. While developing our approaches, we aimed to make approaches agnostic to specific dialects, and given that data exists, we think they might be adaptable to other dialects.

Another aspect that requires consideration is the problem of bias. We identify two main areas of bias in our work: the data we used and our own bias as researchers. In the former case, our work relies on the assumption that annotations are not biased. However, they are made by humans with personal worldviews and biases which, intentionally or not, might have mislabeled the data \cite{sap2022}. Especially for our scenario of AAE texts, human annotators who were not part of the African-American community might have confused dialect use with, e.g., offensiveness \cite{widawski2015}. Such wrong labels, that we assume to be correct, may influence models and evaluations. Similarly, we, the authors of the paper, might have introduced personal biases by applying our specific world views to the problem or unintentionally making false assumptions, leading to potential oversights.

Finally, we conceive that our approaches might be misused in situations where it is helpful for actors to label a product or approach ``debiased.'' Since none of the presented approaches is perfect for detecting social bias or being completely fair for all dialects, as stated above, they are also not ready for production use. Even unintentionally, actors might apply our approaches to their data, wrongly assuming that it identifies all possible cases of social bias. This might, however, rather be an issue regarding the communication of our work to the more general public, as we assume that this will not be problematic for everyone that takes the time to read this paper entirely.

%% file: acl24-dialect-bias-acknowledgements.tex
\section*{Acknowledgments}
\label{sec:acknowledgements}

This work has been supported by the Deutsche Forschungsgemeinschaft (DFG, German Research
Foundation) under project number TRR 318/1 2021 -- 438445824. We thank the anonymous reviewers for their helpful feedback and suggestions.

%% file: acl24-dialect-bias-appendix-a.tex
\section{Qualitative Analysis}
\label{sec:appendix-qualitative-analysis}

\input{table-post-examples}

Selected samples that showcase properties of modeling multiple bias aspects (multitask learning) and/or additionally modeling the AAE dialect simultaneously are shown in Table~\ref{tab:post-examples}. In the remainder of this section, we shortly discuss some of the properties we observed by manually looking at the data and analyzing selected classification results. Each example will be referred to by its index, presented in the first column of the table (i.e., ``Example 2'' refers to the example in the second row).

\subsection{Modeling AAE Improves Latent Concepts}
For selected samples, modeling AAE dialect seems to benefit the offensiveness and lewdness classifications by improving the internal concept representations of the respective aspects. The improved representations seem to help the model abstract from word use towards relying more on contextual information.

The improved representations are shown anecdotally by Examples 4, 11, 14, 22, 23, and 24, all classified as lewd by the \singletask{} model but not by the \singletaskaae{} model. For offensiveness, Examples 1, 2, 3, 10, 12, 15, 17, 18, and 21 showcase the behavior. While all listed examples include words that often indicate offensive or lewd statements, they can also be used outside of said contexts, i.e., in statements that are simply obscene.

As mentioned above, one potential theory might be that, due to the additional AAE dialect modeling and the resulting awareness of dialect language, the model is forced to learn better internal concepts of lewdness and offensiveness less strongly based on word use.

Furthermore, Examples 27 and 28 showcase the positive effect of modeling dialect on the ingroup label. While both examples include the N-word, the use of the AAE dialect version (``a''-ending) and other AAE dialect properties, such as dropping ``have'', suggest that the authors of both texts are ingroup members using ingroup language (this is confirmed by the original annotation). Both examples were classified wrongly by the \singletask{} and \multitask{} models as not ingroup texts, while being classified correctly by the \singletaskaae{} and \multitaskaae{} models.

Note that the listed examples include both, texts with and without properties of AAE dialect. The effect therefore seems to benefit not only texts containing AAE dialect, but also text without AAE dialect. Especially for the lewdness classification, this can also be observed in Table~\ref{tab:results-bias-classification} and Table~\ref{tab:results-per-dialect-bias-classification}. Lastly, modeling AAE can also help to detect offensive statements, which make use of dialect language elements, more reliably (Example 19).

However, since these examples can only anecdotally show this effect, future work might further investigate and attempt to quantify this theory.

It is important to mention that classifications are still not perfect, and selected samples are missed. For instance, Example 19 and 25 both contain AAE dialect elements (such as dropped copula \cite{ziems2022} and use of n-word with an ``a''-ending \cite{rahman2012}) and are not obviously offensive (and are also not labeled as such in the SBIC). However, all evaluated models classified them as offensive, including \multitask{} and \multitaskaae{}.

\subsection{Multitask Learning Improves Label Dependency Modeling}
Due to the conceptual dependency of some labels in the SBIC corpus, certain label-value combinations are nonsensical and should not be predicted. For example, while a text can be intentionally offensive (thus being labeled as \emph{offensive} and \emph{intentional}, such as Example 1), by definition of the aspects, the text cannot be intentional but not offensive. Another example is the label combination of \emph{target group} and \emph{ingroup}: Without referencing a target group in a statement, it is also impossible to state whether the author of the text is part of the referenced group (i.e., the \emph{target group} label is false, while the \emph{ingroup} label is true).

To model and identify this connection, however, it is necessary to consider both aspects simultaneously, as it is lost when considering aspects individually. While the \singletask{} model predicts these impossible combinations for \emph{offensive} and \emph{intentional} only 59 times (e.g., Example 1), both multitask learning variants, \multitask{} and \multitaskaae{}, eliminate the issue and never predict such wrong combinations. While it is not possible to evaluate this behavior on the \emph{target group} and \emph{ingroup} combinations for \singletask{} and \singletaskaae{} (both only predict the majority class for the \emph{ingroup} label), \multitaskaae{} only makes one such error (Example 26). At the same time, \multitask{} never does.

While this error appeared only for a smaller number of samples, the extreme effect observed for the multitask models highlights the benefit of considering multiple bias aspects simultaneously. We assume that the effect is also present in more subtle ways throughout the rest of the dataset. Future work might further investigate this effect.

\subsection{Incorrect Annotations}
In a few cases, it seems that most of the evaluated approaches ``correct for'' wrong annotations in the corpus and predict the, to the authors' perception, correct label value, even though they are mislabeled in the SBIC. Instances of such ``correct misclassifications'' for the \emph{target group} label are Examples 5, 6, 7, 8, 9, and 13.

One possible explanation might be that, for all those instances, the target groups are not within the focus of the original SBIC study presented by \citet{sap2020}. Therefore, these target groups were also not part of the pre-defined options in the annotation interface (shown in the appendix of \citet{sap2020}), and required manual user input. While a reasonable design decision for, what we assume to be the focus of the corpus, it might have caused ``Friction nudges'' \cite{bergram2022}, causing annotators to disregard or consider fewer other target groups.

While such cases are present, they seem to be very few and do not seem to pose a notable challenge, as models can apparently correct for the slight noise.

%% file: table-post-examples.tex
\begin{table*}
    \small
    \renewcommand{\arraystretch}{.95}
    \centering
    \setlength{\tabcolsep}{2pt}

    \begin{tabularx}{\linewidth}{llllllX}
        \toprule
        \bf\# & \bf Offens. & \bf Intent & \bf Lewdn. & \bf Target & \bf Ingroup & \bf Post content \\
        \midrule

        \bf 1 & Yes & Yes & No & No & No & You all make fucking terrible "music" \\
        \midrule

        \bf 2 & No & No & No & No & No & “This guy tko’s people. when I knock people out they don’t fucking move." - eddie alvarez \\
        \midrule

        \bf 3 & No & No & No & No & No & She text me "i hate you, i hate you, i hate you, i hate you" but how the fuck can you hate me? \\
        \midrule

        \bf 4 & No & No & No & No & No & A man already accused of killing two women and suspected in the deaths of as many as 30 more was charged with raping and murdering five other victims. \\
        \midrule

        \bf 5 & Yes & Yes & No & No & No & Always remember, it was normies that enacted the holocaust, acted in the slave trade, and committed every atrocity in history normies are evil \\
        \midrule

        \bf 6 & No & No & No & No & No & Amnesty for 3.6m colonizers would/will be the completion of the demographic coup \\
        \midrule

        \bf 7 & Yes & Yes & No & No & No & Are orphans allowed in a family restaurant. I think you get it. \\
        \midrule

        \bf 8 & No & No & No & No & No & Asking north korea not to make nuclear weapons is a bit like asking an audi driver not to drive too close to the car in front \\
        \midrule

        \bf 9 & No & No & No & No & No & Bpd is a genetic condition caused by having two x chromosomes. When a man is diagnosed with bpd it's just a professional way of saying he's acting like a cunt. \\
        \midrule

        \bf 10 & No & No & No & No & No & Draymond kill me with all that shit he be talking to the refs. Please be quiet boy before you get us all in trouble. \\
        \midrule

        \bf 11 & Yes & Yes & No & Yes & No & Everytime I clean the room my mom tells me "what bitch is coming over this time" lmao what a dick !! \\
        \midrule

        \bf 12 & No & No & No & No & No & gooooooooooaaallll goooooooooooaaalll gooooooooooooaallll pepe pepe pepe!!!! renew his fucking contract!!! \\
        \midrule

        \bf 13 & No & No & No & No & No & I do not blame you, ottawa, like all of our cities, is a multicultural hell \\
        \midrule

        \bf 14 & Yes & No & No & No & No & It's always the people who tell you not to be so obsessed about women and sex who are the ones who can get them anytime they want \\
        \midrule

        \bf 15 & No & No & No & No & No & Natiis ka nya. It means he doesn't fucking care anymore \\
        \midrule

        \bf 16 & No & No & No & No & No & On god I ain't never wanna plex with lil bro them at all. I ain't scared to squabble nobody I just refuse to fight a nigga i had love 4 \\
        \midrule

        \bf 17 & No & No & No & No & No & Bad bitches don't take days off \\
        \midrule

        \bf 18 & No & No & No & No & No & I ain gone lie. Chuckie fucked w| my head a lil bit \\
        \midrule

        \bf 19 & Yes & Yes & No & No & No & Sis don't assume I'm talking about yo nigga and get in yo feelings.. Yo nigga not the only one on my line \\
        \midrule

        \bf 20 & Yes & No & No & No & No & Some of y'all lives don't even be bad.. Y'all jus either ungrateful or just don't know how to put things in perspective \\
        \midrule

        \bf 21 & No & No & No & No & No & Vince once again fucked over undertaker over a blind push. 'taker is over now, and so is me as a fan. \#vincescrewedtaker \\
        \midrule

        \bf 22 & Yes & Yes & No & Yes & No & Why does the myth that women rarely lie about rape persist? \\
        \midrule

        \bf 23 & No & No & No & No & No & Just ordered horny goat weed this shit better fix my ed \\
        \midrule

        \bf 24 & Yes & Yes & No & No & No & Sorry I'm not humble, dick. \\
        \midrule

        \bf 25 & No & No & No & No & No & Your not gonna change a nigga. He is who he is either you accept it or you dont. You cant turn silver into gold. \\
        \midrule

        \bf 26 & No & No & No & No & No & That's some real hood shit, I'll never forget heading out wit my niggas \& realizing that someone stole his gas cap \\
        \midrule

        \bf 27 & Yes & Yes & No & Yes & Yes & I come a city where niggas will hate on you for anything. \\
        \midrule

        \bf 28 & Yes & No & No & Yes & Yes & Nigga Andrew and x been telling me since like 11' everything's for sale and legit everything is for sale you tell me a great number lol \\

        \bottomrule
    \end{tabularx}

    \caption{Example posts from the SBIC test set that highlight different properties and capabilities of the evaluated models. Please note that neither the texts themselves nor the labels assigned to them necessarily reflect the authors opinions. Both are retrieved from the Social Bias Inference Corpus \cite{sap2020}, which collects texts from the internet and annotations from crowdworkers. Some posts were formatted to improve readability.}
    \label{tab:post-examples}
\end{table*}

%% file: acl24-dialect-bias-appendix-b.tex
\section{Further classification results}
\label{sec:appendix-classification-results}

\subsection{AAE Classification}
Table~\ref{tab:results-aae-classification} shows the results of the AAE classification approaches \aaeweightedloss{} and \aaesubsampling{} on the full test set. While the precision values for the positive class seem small, it is to be considered that this is a needle-in-the-haystack problem: Only about 2\% of the test cases are positive, meaning that the best result (.07 of our approach \aaeweightedloss{}) is 3.5 times better than guessing. An increase of two points over the TwitterAAE baseline (.07 vs.\ .05) also indicates a notable learning effect, classifying about 40\% more correctly.

\input{table-results-aae-classification}

\subsection{Social Bias Classification}
Table~\ref{tab:results-bias-classification-negative-class} and Table~\ref{tab:results-bias-classification-macro} show the negative class and macro averaged F$_1$-scores of the social bias classification. Results for the positive class are reported in Table~\ref{tab:results-bias-classification}, as part of Section~\ref{sec:results}.

\input{table-results-bias-classification-negative-class}

\input{table-results-bias-classification-macro}

\subsection{Encoder model}
To verify that our results are not specific to the chosen DeBERTa-v3-base encoder model, Table~\ref{tab:results-bias-classification-positive-class-roberta} shows the results for the positive F$_1$-scores, matching Table~\ref{tab:results-bias-classification}. The results highlight that, while the RoBERTa-base models do not seem to benefit to the same degree the DeBERTa-v3-base models do, most advantages of the multitask learning setup and dialect modeling discussed in Section~\ref{sec:results} also hold in this setup.

\input{table-results-bias-classification-positive-class-roberta}

\subsection{Social Bias Classification Scores per Dialect}
Table~\ref{tab:results-per-dialect-bias-classification-offensive}, Table~\ref{tab:results-per-dialect-bias-classification-intent}, Table~\ref{tab:results-per-dialect-bias-classification-lewd}, Table~\ref{tab:results-per-dialect-bias-classification-group} and Table~\ref{tab:results-per-dialect-bias-classification-ingroup} show the per-class precision, recall and F$_1$-scores for the social bias classifications. Macro-averaged F$_1$ scores are reported in Table~\ref{tab:results-per-dialect-bias-classification}, as part of Section~\ref{sec:results}.

\input{table-results-per-dialect-bias-classification-offensive}

\input{table-results-per-dialect-bias-classification-intent}

\input{table-results-per-dialect-bias-classification-lewd}

\input{table-results-per-dialect-bias-classification-group}

\input{table-results-per-dialect-bias-classification-ingroup}

\subsection{Fairness in Social Bias Classification}
Table~\ref{tab:results-fairness-intent}, Table~\ref{tab:results-fairness-lewd}, Table~\ref{tab:results-fairness-group}, and Table~\ref{tab:results-fairness-ingroup} show True Positive Rate, False Positive Rate and Positive Predictive Value scores on the positive class for the bias aspects \textit{intent}, \textit{lewdness}, \textit{target group}, and \textit{in-group}, respectively. Results for the \textit{offensiveness} aspect are reported in Table~\ref{tab:results-fairness-offensive}, as part of  Section~\ref{sec:results}.

\input{table-results-fairness-intent}

\input{table-results-fairness-lewd}

\input{table-results-fairness-group}

\input{table-results-fairness-ingroup}

%% file: table-results-aae-classification.tex
\begin{table}
    \small
    \centering
    \setlength{\tabcolsep}{2pt}

    \begin{tabular}{lrrrrrrrrr}
        \toprule
        & \multicolumn{3}{c}{\textbf{Precision ↑}} & \multicolumn{3}{c}{\textbf{Recall ↑}} & \multicolumn{3}{c}{\textbf{F$_1$ ↑}}  \\
        \cmidrule(l@{3pt}r@{3pt}){2-4}\cmidrule(l@{3pt}r@{3pt}){5-7}\cmidrule(l@{3pt}r@{3pt}){8-10}
        \textbf{Model} & \textbf{Pos} & \textbf{Neg} & \textbf{Mac} & \textbf{Pos} & \textbf{Neg} & \textbf{Mac} & \textbf{Pos} & \textbf{Neg} & \textbf{Mac} \\
        \midrule

        Majority & .00 & .98 & .49 & .00 & \bf 1.0 & .50 & .00 & \bf .99 & .50 \\

        Random & .02 & .98 & .50 & .50 & .50 & .50 & .04 & .66 & .35 \\
        [.5em]
        TwitterAAE & .05 & \textbf{.99} & .52 & .64 & .77 & .70 & .10 & .87 & .48 \\
        [.5em]

        \aaeweightedloss & \textbf{.07} & \textbf{.99} & \textbf{.53} & .74 & .81 & \textbf{.78} & \textbf{.13} & {.89} & \textbf{.51} \\

        \aaesubsampling & .06 & \textbf{.99} & \textbf{.53} & \textbf{.78} & .77 & .77 & .11 & .86 & .49 \\

        \bottomrule
    \end{tabular}

    \caption{African-American English dialect classification results on the full test set: Postive class (\textit{Pos}), negative class (\textit{Neg}) and macro averaged (\textit{Mac}). Bold values highlights the best result in each column. While \aaesubsampling{} seems better in finding dialect texts (higher recall for the positive class), \aaeweightedloss{} performs notably better overall.}
    \label{tab:results-aae-classification}
\end{table}

%% file: table-results-bias-classification-negative-class.tex
\begin{table}
    \small
    \centering
    \setlength{\tabcolsep}{2.7pt}

    \begin{tabular}{l@{\hspace*{-1em}}rrrrr}
        \toprule
        \textbf{Model} & \textbf{Offens.} & \textbf{Intent} & \textbf{Lewdn.} & \textbf{Target} & \textbf{Ingroup} \\
        \midrule
        Majority & .000 & .000 & .949 & .741 & \bf .990 \\
        Random & .463 & .479 & .639 & .543 & .654 \\
        [.5em]
        GPT-2 & -- & -- & -- & -- & -- \\
        Few-shot & -- & -- & -- & -- & -- \\
        [.5em]
        \singletask & .820 & \bf .834 & .973 & .869 & \bf .990 \\
        \singletaskaae & .820 & .833 & .974 & \bf .876 & \bf .990 \\
        [.5em]
        \multitask & \bf .830 & .833 & \bf .975 & \bf .876 & .989 \\
        \multitaskaae & .824 & .832 & .974 & .874 & \bf .990 \\

        \bottomrule
    \end{tabular}

    \caption{Bias classification results (negative class F$_1$-scores, averaged over five random seeds) for each aspect: offensiveness, intent, lewdness, target group, and ingroup.}
    \label{tab:results-bias-classification-negative-class}
\end{table}

%% file: table-results-bias-classification-macro.tex
\begin{table}
    \small
    \centering
    \setlength{\tabcolsep}{2.7pt}

    \begin{tabular}{l@{\hspace*{-1em}}rrrrr}
        \toprule
        \textbf{Model} & \textbf{Offens.} & \textbf{Intent} & \textbf{Lewdn.} & \textbf{Target} & \textbf{Ingroup} \\
        \midrule
        Majority & .366 & .347 & .475 & .370 & .495 \\
        Random & .496 & .492 & .402 & .499 & .346 \\
        [.5em]
        GPT-2 & -- & -- & -- & -- & -- \\
        Few-shot & -- & -- & -- & -- & -- \\
        [.5em]
        \singletask & .848 & .847 & .859 & .851 & .495 \\
        \singletaskaae & .847 & .847 & .864 & .854 & .549 \\
        [.5em]
        \multitask & \bf .856 & \bf .849 & \bf .866 & \bf .854 & \bf .612 \\
        \multitaskaae & .852 & .848 & .863 & .853 & .608 \\

        \bottomrule
    \end{tabular}

    \caption{Bias classification results (macro F$_1$-scores, averaged over five random seeds) for each aspect: offensiveness, intent, lewdness, target group, and ingroup.}
    \label{tab:results-bias-classification-macro}
\end{table}

%% file: table-results-bias-classification-positive-class-roberta.tex
\begin{table}[t]
    \small
    \renewcommand{\arraystretch}{1}
    \centering
    \setlength{\tabcolsep}{2.7pt}
    \begin{tabular}{l@{\hspace*{-1em}}rrrrr}
        \toprule
        \textbf{Model} & \textbf{Offens.} & \textbf{Intent} & \textbf{Lewdn.} & \textbf{Target} & \textbf{Ingroup} \\
        \midrule
        Majority & .732 & .694 & .000 & .000 & .000 \\
        Random & .529 & .504 & .165 & .456 & .038 \\
        [.5em]
        GPT-2 & .788 & .786 & \bf .807 & .699 & .000 \\
        Few-shot & .822 & .798 & .411 & .737 & -- \\
        [.5em]
        \singletask & .875 & \bf .859 & .722 & .826 & .000 \\
        \singletaskaae & .875 & .857 & .752 & \bf .833 & .281 \\
        [.5em]
        \multitask & .875 & .857 & .754 & .826 & \bf .290 \\
        \multitaskaae & \bf .877 & \bf .859 & .751 & .828 & .213 \\
        \bottomrule
    \end{tabular}
    \caption{Bias classification results (positive-class F$_1$, averaged over five random seeds) for each aspect: offensiveness, intent, lewdness, target group, and ingroup. Results obtained using a RoBERTa-base encoder model, as comapred to Table~\ref{tab:results-bias-classification}, where DeBERTa-v3-base is used as encoder model.}
    \label{tab:results-bias-classification-positive-class-roberta}
\end{table}

%% file: table-results-per-dialect-bias-classification-offensive.tex
\begin{table*}
    \small
    \centering
    \setlength{\tabcolsep}{2.5pt}

    \begin{tabular}{lrrrrrrrrrrrrrrrrrr}
        \toprule
        \bf Offensiveness & \multicolumn{6}{c}{\textbf{Precision}} & \multicolumn{6}{c}{\textbf{Recall}} & \multicolumn{6}{c}{\textbf{F$_1$}} \\
        \cmidrule(l@{2pt}r@{2pt}){2-7}\cmidrule(l@{2pt}r@{2pt}){8-13}\cmidrule(l@{2pt}r@{2pt}){14-19}
        & \multicolumn{2}{c}{\textbf{Positive}} & \multicolumn{2}{c}{\textbf{Negative}} & \multicolumn{2}{c}{\textbf{Macro}} & \multicolumn{2}{c}{\textbf{Positive}} & \multicolumn{2}{c}{\textbf{Negative}} & \multicolumn{2}{c}{\textbf{Macro}} & \multicolumn{2}{c}{\textbf{Positive}} & \multicolumn{2}{c}{\textbf{Negative}} & \multicolumn{2}{c}{\textbf{Macro}}  \\
        \cmidrule(l@{2pt}r@{2pt}){2-3}\cmidrule(l@{2pt}r@{2pt}){4-5}\cmidrule(l@{2pt}r@{2pt}){6-7}\cmidrule(l@{2pt}r@{2pt}){8-9}\cmidrule(l@{2pt}r@{2pt}){10-11}\cmidrule(l@{2pt}r@{2pt}){12-13}\cmidrule(l@{2pt}r@{2pt}){14-15}\cmidrule(l@{2pt}r@{2pt}){16-17}\cmidrule(l@{2pt}r@{2pt}){18-19}
        \textbf{Model} & \textbf{NA.} & \textbf{AA.} & \textbf{NA.} & \textbf{AA.} & \textbf{NA.} & \textbf{AA.} & \textbf{NA.} & \textbf{AA.} & \textbf{NA.} & \textbf{AA.} & \textbf{NA.} & \textbf{AA.} & \textbf{NA.} & \textbf{AA.} & \textbf{NA.} & \textbf{AA.} & \textbf{NA.} & \textbf{AA.} \\
        \midrule
        Majority & .566 & .658 & .000 & .000 & .283 & .329 & \bf 1.00 & \bf 1.00 & .000 & .000 & .500 & .500 & .723 & .794 & .000 & .000 & .361 & .397 \\
        Random &  .564 & .670 & .432 & .354 & .498 & .512 & .485 & .514 & .510 & .513 & .498 & .513 & .522 & .582 & .468 & .419 & .495 & .500 \\
        [.5em]
        \singletask & .860 & .826 & .853 & .800 & .857 & .813 & .893 & .918 & .810 & .628 & .852 & .773 & .876 & .870 & .831 & .704 & .854 & .787 \\
        \singletaskaae & .857 & .836 & .851 & \bf .844 & .854 & .840 & .891 & .938 & .807 & .646 & .849 & .792 & .874 & .884 & .828 & .732 & .851 & .808 \\
        [.5em]
        \multitask & \bf .866 & \bf .845 & \bf .859 & .842 & \bf .862 & \bf .843 & .896 & .934 & \bf .819 & \bf .669 & \bf .858 & \bf .802 & \bf .881 & \bf .887 & \bf .838 & \bf .745 & \bf .860 & \bf .816 \\
        \multitaskaae & .861 & .836 & .857 & .838 & .859 & .837 & .896 & .935 & .811 & .647 & .854 & .791 & .878 & .883 & .833 & .730 & .856 & .806 \\
        \bottomrule
    \end{tabular}

    \caption{Offensiveness classification results on texts written with (\textit{AA.}) and without (\textit{NA.}) AAE dialect for the positive (\textit{Positive}) and negative class (\textit{Negative}), and macro averaged (\textit{Macro}). All scores are averaged over five random seeds. Bold indicates best results per column.}
    \label{tab:results-per-dialect-bias-classification-offensive}
\end{table*}

%% file: table-results-per-dialect-bias-classification-intent.tex
\begin{table*}
    \small
    \centering
    \setlength{\tabcolsep}{2.5pt}

    \begin{tabular}{lrrrrrrrrrrrrrrrrrr}
        \toprule
        \bf Intent & \multicolumn{6}{c}{\textbf{Precision}} & \multicolumn{6}{c}{\textbf{Recall}} & \multicolumn{6}{c}{\textbf{F$_1$}} \\
        \cmidrule(l@{2pt}r@{2pt}){2-7}\cmidrule(l@{2pt}r@{2pt}){8-13}\cmidrule(l@{2pt}r@{2pt}){14-19}
        & \multicolumn{2}{c}{\textbf{Positive}} & \multicolumn{2}{c}{\textbf{Negative}} & \multicolumn{2}{c}{\textbf{Macro}} & \multicolumn{2}{c}{\textbf{Positive}} & \multicolumn{2}{c}{\textbf{Negative}} & \multicolumn{2}{c}{\textbf{Macro}} & \multicolumn{2}{c}{\textbf{Positive}} & \multicolumn{2}{c}{\textbf{Negative}} & \multicolumn{2}{c}{\textbf{Macro}}  \\
        \cmidrule(l@{2pt}r@{2pt}){2-3}\cmidrule(l@{2pt}r@{2pt}){4-5}\cmidrule(l@{2pt}r@{2pt}){6-7}\cmidrule(l@{2pt}r@{2pt}){8-9}\cmidrule(l@{2pt}r@{2pt}){10-11}\cmidrule(l@{2pt}r@{2pt}){12-13}\cmidrule(l@{2pt}r@{2pt}){14-15}\cmidrule(l@{2pt}r@{2pt}){16-17}\cmidrule(l@{2pt}r@{2pt}){18-19}
        \textbf{Model} & \textbf{NA.} & \textbf{AA.} & \textbf{NA.} & \textbf{AA.} & \textbf{NA.} & \textbf{AA.} & \textbf{NA.} & \textbf{AA.} & \textbf{NA.} & \textbf{AA.} & \textbf{NA.} & \textbf{AA.} & \textbf{NA.} & \textbf{AA.} & \textbf{NA.} & \textbf{AA.} & \textbf{NA.} & \textbf{AA.} \\
        \midrule
        Majority & .525 & .583 & .000 & .000 & .262 & .292 & \bf 1.00 & \bf 1.00 & .000 & .000 & .500 & .500 & .688 & .737 & .000 & .000 & .344 & .368 \\
        Random & .517 & .577 & .468 & .411 & .493 & .494 & .490 & .460 & .496 & .528 & .493 & .494 & .503 & .512 & .482 & .462 & .492 & .487 \\
        [.5em]
        \singletask & \bf .851 & \bf .783 & .859 & .837 & .855 & .810 & .877 & .910 & \bf .831 & \bf .648 & .854 & .779 & .864 & .842 & \bf .845 & .731 & .854 & .786 \\
        \singletaskaae & .847 & .781 & .862 & \bf .862 & .855 & \bf .822 & .880 & .927 & .825 & .637 & .853 & \bf .782 & .863 & \bf .848 & .843 & \bf .732 & .853 & \bf .790 \\
        [.5em]
        \multitask & .843 & .777 & \bf .875 & .856 & \bf .859 & .816 & .894 & .924 & .816 & .628 & \bf .855 & .776 & \bf .868 & .844 & .844 & .724 & \bf .856 & .784 \\
        \multitaskaae & .842 & .776 & .874 & .856 & .858 & .816 & .894 & .925 & .814 & .626 & .854 & .775 & .867 & .844 & .843 & .723 & .855 & .783 \\
        \bottomrule
    \end{tabular}

    \caption{Intent classification results on texts written with (\textit{AA.}) and without (\textit{NA.}) AAE dialect for the positive (\textit{Positive}) and negative class (\textit{Negative}), and macro averaged (\textit{Macro}). All scores are averaged over five random seeds. Bold indicates best results per column.}
    \label{tab:results-per-dialect-bias-classification-intent}
\end{table*}

%% file: table-results-per-dialect-bias-classification-lewd.tex
\begin{table*}
    \small
    \centering
    \setlength{\tabcolsep}{2.5pt}

    \begin{tabular}{lrrrrrrrrrrrrrrrrrr}
        \toprule
        \bf Lewdness & \multicolumn{6}{c}{\textbf{Precision}} & \multicolumn{6}{c}{\textbf{Recall}} & \multicolumn{6}{c}{\textbf{F$_1$}} \\
        \cmidrule(l@{2pt}r@{2pt}){2-7}\cmidrule(l@{2pt}r@{2pt}){8-13}\cmidrule(l@{2pt}r@{2pt}){14-19}
        & \multicolumn{2}{c}{\textbf{Positive}} & \multicolumn{2}{c}{\textbf{Negative}} & \multicolumn{2}{c}{\textbf{Macro}} & \multicolumn{2}{c}{\textbf{Positive}} & \multicolumn{2}{c}{\textbf{Negative}} & \multicolumn{2}{c}{\textbf{Macro}} & \multicolumn{2}{c}{\textbf{Positive}} & \multicolumn{2}{c}{\textbf{Negative}} & \multicolumn{2}{c}{\textbf{Macro}}  \\
        \cmidrule(l@{2pt}r@{2pt}){2-3}\cmidrule(l@{2pt}r@{2pt}){4-5}\cmidrule(l@{2pt}r@{2pt}){6-7}\cmidrule(l@{2pt}r@{2pt}){8-9}\cmidrule(l@{2pt}r@{2pt}){10-11}\cmidrule(l@{2pt}r@{2pt}){12-13}\cmidrule(l@{2pt}r@{2pt}){14-15}\cmidrule(l@{2pt}r@{2pt}){16-17}\cmidrule(l@{2pt}r@{2pt}){18-19}
        \textbf{Model} & \textbf{NA.} & \textbf{AA.} & \textbf{NA.} & \textbf{AA.} & \textbf{NA.} & \textbf{AA.} & \textbf{NA.} & \textbf{AA.} & \textbf{NA.} & \textbf{AA.} & \textbf{NA.} & \textbf{AA.} & \textbf{NA.} & \textbf{AA.} & \textbf{NA.} & \textbf{AA.} & \textbf{NA.} & \textbf{AA.} \\
        \midrule
        Majority & .000 & .000 & .907 & .877 & .454 & .438 & .000 & .000 & \bf 1.00 & \bf 1.00 & .500 & .500 & .000 & .000 & .951 & .934 & .476 & .467 \\
        Random & .092 & .143 & .907 & .895 & .500 & .519 & .509 & .565 & .490 & .522 & .500 & .544 & .156 & .228 & .636 & .660 & .396 & .444 \\
        [.5em]
        \singletask & .757 & \bf .783 & .974 & .954 & .865 & \bf .869 & .739 & .670 & .976 & .973 & .858 & .822 & .748 & .721 & .975 & \bf .964 & .861 & .842 \\
        \singletaskaae & .756 & .748 & \bf .976 & .955 & .866 & .852 & \bf .768 & .678 & .975 & .968 & \bf .871 & .823 & .762 & .711 & \bf .976 & .961 & .869 & .836 \\
        [.5em]
        \multitask & \bf .779 & .760 & .975 & .955 & \bf .877 & .858 & .749 & .678 & .978 & .970 & .864 & .824 & \bf .764 & .717 & \bf .976 & .963 & \bf .870 & .840 \\
        \multitaskaae & .771 & .781 & .974 & \bf .956 & .872 & .868 & .740 & \bf .681 & .978 & .973 & .859 & \bf .827 & .755 & \bf .728 & \bf .976 & \bf .964 & .865 & \bf .846 \\
        \bottomrule
    \end{tabular}

    \caption{Lewdness classification results on texts written with (\textit{AA.}) and without (\textit{NA.}) AAE dialect for the positive (\textit{Positive}) and negative class (\textit{Negative}), and macro averaged (\textit{Macro}). All scores are averaged over five random seeds. Bold indicates best results per column.}
    \label{tab:results-per-dialect-bias-classification-lewd}
\end{table*}

%% file: table-results-per-dialect-bias-classification-group.tex
\begin{table*}
    \small
    \centering
    \setlength{\tabcolsep}{2.5pt}

    \begin{tabular}{lrrrrrrrrrrrrrrrrrr}
        \toprule
        \bf Group & \multicolumn{6}{c}{\textbf{Precision}} & \multicolumn{6}{c}{\textbf{Recall}} & \multicolumn{6}{c}{\textbf{F$_1$}} \\
        \cmidrule(l@{2pt}r@{2pt}){2-7}\cmidrule(l@{2pt}r@{2pt}){8-13}\cmidrule(l@{2pt}r@{2pt}){14-19}
        & \multicolumn{2}{c}{\textbf{Positive}} & \multicolumn{2}{c}{\textbf{Negative}} & \multicolumn{2}{c}{\textbf{Macro}} & \multicolumn{2}{c}{\textbf{Positive}} & \multicolumn{2}{c}{\textbf{Negative}} & \multicolumn{2}{c}{\textbf{Macro}} & \multicolumn{2}{c}{\textbf{Positive}} & \multicolumn{2}{c}{\textbf{Negative}} & \multicolumn{2}{c}{\textbf{Macro}}  \\
        \cmidrule(l@{2pt}r@{2pt}){2-3}\cmidrule(l@{2pt}r@{2pt}){4-5}\cmidrule(l@{2pt}r@{2pt}){6-7}\cmidrule(l@{2pt}r@{2pt}){8-9}\cmidrule(l@{2pt}r@{2pt}){10-11}\cmidrule(l@{2pt}r@{2pt}){12-13}\cmidrule(l@{2pt}r@{2pt}){14-15}\cmidrule(l@{2pt}r@{2pt}){16-17}\cmidrule(l@{2pt}r@{2pt}){18-19}
        \textbf{Model} & \textbf{NA.} & \textbf{AA.} & \textbf{NA.} & \textbf{AA.} & \textbf{NA.} & \textbf{AA.} & \textbf{NA.} & \textbf{AA.} & \textbf{NA.} & \textbf{AA.} & \textbf{NA.} & \textbf{AA.} & \textbf{NA.} & \textbf{AA.} & \textbf{NA.} & \textbf{AA.} & \textbf{NA.} & \textbf{AA.} \\
        \midrule
        Majority & .000 & .000 & .592 & .562 & .296 & .281 & .000 & .000 & \bf 1.00 & \bf 1.00 & .500 & .500 & .000 & .000 & .743 & .719 & .372 & .360 \\
        Random & .412 & .440 & .595 & .564 & .503 & .502 & .504 & .510 & .503 & .494 & .503 & .502 & .453 & .473 & .545 & .526 & .499 & .499 \\
        [.5em]
        \singletask & .806 & .674 & \bf .916 & \bf .862 & .861 & \bf .768 & \bf .887 & \bf .859 & .852 & .674 & .869 & \bf .766 & .844 & \bf .754 & .883 & .755 & .863 & .755 \\
        \singletaskaae & .821 & .699 & .908 & .824 & \bf .865 & .761 & .873 & .800 & .869 & .731 & \bf .871 & .765 & \bf .846 & .746 & .888 & \bf .774 & \bf .867 & \bf .760 \\
        [.5em]
        \multitask & \bf .826 & \bf .700 & .905 & .814 & \bf .865 & .757 & .866 & .784 & .874 & .737 & .870 & .760 & .845 & .739 & \bf .889 & .773 & \bf .867 & .756 \\
        \multitaskaae & .822 & .687 & .908 & .813 & \bf .865 & .750 & .872 & .788 & .869 & .720 & \bf .871 & .754 & \bf .846 & .734 & .888 & .764 & \bf .867 & .749 \\
        \bottomrule
    \end{tabular}

    \caption{Group classification results on texts written with (\textit{AA.}) and without (\textit{NA.}) AAE dialect for the positive (\textit{Positive}) and negative class (\textit{Negative}), and macro averaged (\textit{Macro}). All scores are averaged over five random seeds. Bold indicates best results per column.}
    \label{tab:results-per-dialect-bias-classification-group}
\end{table*}

%% file: table-results-per-dialect-bias-classification-ingroup.tex
\begin{table*}
    \small
    \centering
    \setlength{\tabcolsep}{2.5pt}

    \begin{tabular}{lrrrrrrrrrrrrrrrrrr}
        \toprule
        \bf In-group & \multicolumn{6}{c}{\textbf{Precision}} & \multicolumn{6}{c}{\textbf{Recall}} & \multicolumn{6}{c}{\textbf{F$_1$}} \\
        \cmidrule(l@{2pt}r@{2pt}){2-7}\cmidrule(l@{2pt}r@{2pt}){8-13}\cmidrule(l@{2pt}r@{2pt}){14-19}
        & \multicolumn{2}{c}{\textbf{Positive}} & \multicolumn{2}{c}{\textbf{Negative}} & \multicolumn{2}{c}{\textbf{Macro}} & \multicolumn{2}{c}{\textbf{Positive}} & \multicolumn{2}{c}{\textbf{Negative}} & \multicolumn{2}{c}{\textbf{Macro}} & \multicolumn{2}{c}{\textbf{Positive}} & \multicolumn{2}{c}{\textbf{Negative}} & \multicolumn{2}{c}{\textbf{Macro}}  \\
        \cmidrule(l@{2pt}r@{2pt}){2-3}\cmidrule(l@{2pt}r@{2pt}){4-5}\cmidrule(l@{2pt}r@{2pt}){6-7}\cmidrule(l@{2pt}r@{2pt}){8-9}\cmidrule(l@{2pt}r@{2pt}){10-11}\cmidrule(l@{2pt}r@{2pt}){12-13}\cmidrule(l@{2pt}r@{2pt}){14-15}\cmidrule(l@{2pt}r@{2pt}){16-17}\cmidrule(l@{2pt}r@{2pt}){18-19}
        \textbf{Model} & \textbf{NA.} & \textbf{AA.} & \textbf{NA.} & \textbf{AA.} & \textbf{NA.} & \textbf{AA.} & \textbf{NA.} & \textbf{AA.} & \textbf{NA.} & \textbf{AA.} & \textbf{NA.} & \textbf{AA.} & \textbf{NA.} & \textbf{AA.} & \textbf{NA.} & \textbf{AA.} & \textbf{NA.} & \textbf{AA.} \\
        \midrule
        Majority & .000 & .000 & .988 & .932 & .494 & .466 & .000 & .000 & \bf 1.00 & \bf 1.00 & .500 & .500 & .000 & .000 & \bf .994 & \bf .965 & .497 & .482 \\
        Random & .014 & .062 & \bf .989 & .926 & .501 & .494 & \bf .569 & \bf .474 & .492 & .478 & .530 & .476 & .027 & .110 & .657 & .630 & .342 & .370 \\
        [.5em]
        \singletask & .000 & .000 & .988 & .932 & .494 & .466 & .000 & .000 & .000 & .000 & .500 & .500 & .000 & .000 & \bf .994 & \bf .965 & .497 & .482 \\
        \singletaskaae & .233 & .140 & .988 & .941 & .611 & .540 & .039 & .147 & .000 & .980 & .519 & .564 & .066 & .141 & \bf .994 & .960 & .530 & .550 \\
        [.5em]
        \multitask & .578 & .331 & \bf .989 & \bf .949 & .783 & .640 & .082 & .295 & .999 & .955 & \bf .541 & \bf .625 & \bf .143 & .309 & \bf .994 & .952 & \bf .569 & .630 \\
        \multitaskaae & \bf .653 & \bf .375 & .988 & \bf .949 & \bf .821 & \bf .662 & .063 & .284 & .999 & .966 & .531 & \bf .625 & .113 & \bf .322 & \bf .994 & .957 & .553 & \bf .639 \\
        \bottomrule
    \end{tabular}

    \caption{In-group classification results on texts written with (\textit{AA.}) and without (\textit{NA.}) AAE dialect for the positive (\textit{Positive}) and negative class (\textit{Negative}), and macro averaged (\textit{Macro}). All scores are averaged over five random seeds. Bold indicates best results per column.}
    \label{tab:results-per-dialect-bias-classification-ingroup}
\end{table*}

%% file: table-results-fairness-intent.tex
\begin{table*}
    \small
    \centering

    \begin{tabular}{lrrrrrrrrr}
        \toprule

        \bf Intent & \multicolumn{3}{c}{\textbf{True Positive Rate ↑}} & \multicolumn{3}{c}{\textbf{False Positive Rate ↓}} & \multicolumn{3}{c}{\textbf{Positive Predictive Value ↑}} \\

        \cmidrule(l@{5pt}r@{5pt}){2-4}\cmidrule(l@{5pt}r@{5pt}){5-7}\cmidrule(l@{5pt}r@{5pt}){8-10}
        \textbf{Model} & \textbf{$\lnot$AAE} & \textbf{AAE} & \textbf{Delta} & \textbf{$\lnot$AAE} & \textbf{AAE} & \textbf{Delta} & \textbf{$\lnot$AAE} & \textbf{AAE} & \textbf{Delta} \\

        \midrule

        \singletask    & .877 & .910 & .033 & \bf .169 & \bf .352 & \bf .183 & \bf.851 & \bf.783 & .068 \\
        \singletaskaae & .880 & \bf .927 & .047 & .175 & .363 & .188 & .847 & .781 & \bf.066 \\
        [.5em]
        \multitask     & \bf .894 & .924 & \bf .030 & .184 & .372 & .188 & .843 & .777 & \bf.066 \\
        \multitaskaae  & \bf .894 & .925 & .031 & .186 & .374 & .188 & .842 & .776 & \bf.066 \\

        \bottomrule
    \end{tabular}

    \caption{Results for the \textit{intent} aspect per approach for True Positive Rates, False Positive Rates, and Positive Predictive Value, the elements of the fairness metrics \textit{Equalized odds} and \textit{Predictive parity}. All scores are averaged over five random seeds. Bold indicates best results in each column, arrows indicate whether higher (↑) or lower (↓) scores are better.}
    \label{tab:results-fairness-intent}
\end{table*}

%% file: table-results-fairness-lewd.tex
\begin{table*}
    \small
    \centering

    \begin{tabular}{lrrrrrrrrr}
        \toprule

        \bf Lewdness & \multicolumn{3}{c}{\textbf{True Positive Rate ↑}} & \multicolumn{3}{c}{\textbf{False Positive Rate ↓}} & \multicolumn{3}{c}{\textbf{Positive Predictive Value ↑}} \\

        \cmidrule(l@{5pt}r@{5pt}){2-4}\cmidrule(l@{5pt}r@{5pt}){5-7}\cmidrule(l@{5pt}r@{5pt}){8-10}
        \textbf{Model} & \textbf{$\lnot$AAE} & \textbf{AAE} & \textbf{Delta} & \textbf{$\lnot$AAE} & \textbf{AAE} & \textbf{Delta} & \textbf{$\lnot$AAE} & \textbf{AAE} & \textbf{Delta} \\

        \midrule

        \singletask & .739 & .670 & .069 & .024 & \bf .027 & \bf .003 & .757 & \bf .783 & .026 \\
        \singletaskaae & \bf .768 & .678 & .090 & .025 & .032 & .007 & .756 & .748 & \bf .008 \\
        [.5em]
        \multitask & .749 & .678 & .071 & \bf .022 & .030 & .008 & \bf .779 & .760 & .019 \\
        \multitaskaae & .740 & \bf .681 & \bf .059 & \bf .022 & \bf .027 & .005 & .771 & .781 & .010 \\

        \bottomrule
    \end{tabular}

    \caption{Results for the \textit{lewdness} aspect per approach for True Positive Rates, False Positive Rates, and Positive Predictive Value, the elements of the fairness metrics \textit{Equalized odds} and \textit{Predictive parity}. All scores are averaged over five random seeds. Bold indicates best results in each column, arrows indicate whether higher (↑) or lower (↓) scores are better.}
    \label{tab:results-fairness-lewd}
\end{table*}

%% file: table-results-fairness-group.tex
\begin{table*}
    \small
    \centering

    \begin{tabular}{lrrrrrrrrr}
        \toprule

        \bf Group & \multicolumn{3}{c}{\textbf{True Positive Rate ↑}} & \multicolumn{3}{c}{\textbf{False Positive Rate ↓}} & \multicolumn{3}{c}{\textbf{Positive Predictive Value ↑}} \\

        \cmidrule(l@{5pt}r@{5pt}){2-4}\cmidrule(l@{5pt}r@{5pt}){5-7}\cmidrule(l@{5pt}r@{5pt}){8-10}
        \textbf{Model} & \textbf{$\lnot$AAE} & \textbf{AAE} & \textbf{Delta} & \textbf{$\lnot$AAE} & \textbf{AAE} & \textbf{Delta} & \textbf{$\lnot$AAE} & \textbf{AAE} & \textbf{Delta} \\

        \midrule

       \singletask & \bf .887 & \bf .859 & \bf .028 & .148 & .326 & .178 & .806 & .674 & .132 \\
       \singletaskaae & .873 & .800 & .073 & .131 & .269 & .138 & .821 & .699 & \bf .122 \\
       [.5em]
       \multitask & .866 & .784 & .082 & \bf .126 & \bf .263 & \bf .137 & \bf .826 & \bf .700 & .126 \\
       \multitaskaae & .872 & .788 & .084 & .131 & .280 & .149 & .822 & .687 & .135 \\

        \bottomrule
    \end{tabular}

    \caption{Results for the \textit{group} aspect per approach for True Positive Rates, False Positive Rates, and Positive Predictive Value, the elements of the fairness metrics \textit{Equalized odds} and \textit{Predictive parity}. All scores are averaged over five random seeds. Bold indicates best results in each column, arrows indicate whether higher (↑) or lower (↓) scores are better.}
    \label{tab:results-fairness-group}
\end{table*}

%% file: table-results-fairness-ingroup.tex
\begin{table*}
    \small
    \centering

    \begin{tabular}{lrrrrrrrrr}
        \toprule

        \bf In-group & \multicolumn{3}{c}{\textbf{True Positive Rate ↑}} & \multicolumn{3}{c}{\textbf{False Positive Rate ↓}} & \multicolumn{3}{c}{\textbf{Positive Predictive Value ↑}} \\

        \cmidrule(l@{5pt}r@{5pt}){2-4}\cmidrule(l@{5pt}r@{5pt}){5-7}\cmidrule(l@{5pt}r@{5pt}){8-10}
        \textbf{Model} & \textbf{$\lnot$AAE} & \textbf{AAE} & \textbf{Delta} & \textbf{$\lnot$AAE} & \textbf{AAE} & \textbf{Delta} & \textbf{$\lnot$AAE} & \textbf{AAE} & \textbf{Delta} \\

        \midrule

        \singletask & .000 & .000 & \bf .000 & \bf .000 & \bf .000 & \bf .000 & .000 & .000 & \bf .000 \\
        \singletaskaae & .039 & .147 & .108 & \bf .000 & .020 & .020 & .233 & .140 & .093 \\
        [.5em]
        \multitask & \bf .082 & \bf .295 & .213 & .001 & .045 & .044 & .578 & .331 & .247 \\
        \multitaskaae & .063 & .284 & .221 & .001 & .034 & .033 & \bf .653 & \bf .375 & .278 \\

        \bottomrule
    \end{tabular}

    \caption{Results for the \textit{in-group} aspect per approach for True Positive Rates, False Positive Rates, and Positive Predictive Value, the elements of the fairness metrics \textit{Equalized odds} and \textit{Predictive parity}. All scores are averaged over five random seeds. Bold indicates best results in each column, arrows indicate whether higher (↑) or lower (↓) scores are better.}
    \label{tab:results-fairness-ingroup}
\end{table*}

%% file: acl24-dialect-bias-appendix-c.tex
\section{Experimental Details}
\label{sec:appendix-experimental-details}

\subsection{AAE Classification}
Models for the AAE classification (\aaeweightedloss{} and \aaesubsampling{}) were fine-tuned for three epochs on three A100-SXM4-80GB GPUs and a batch size of 270. To keep training time reasonable, given the size of the dataset, we fine-tune the model with bf16 mixed precision using the DeepSpeed \cite{rajbhandari2020} integration of the Huggingface library \cite{wolf2020}. With this setup, fine-tuning takes around 70 hours for \aaeweightedloss{}, and around 17 hours for \aaesubsampling{}.

For all models, we report results for a single training and inference run.

\subsection{Social Bias Classification}
The \singletask{} model for the social bias classification was fine-tuned for three epochs on two A100-SXM4-80GB GPUs, using a batch size of 64. To increase training speed, we fine-tune the model using the DeepSpeed \cite{rajbhandari2020} integration of the Huggingface library \cite{wolf2020}. With this setup, fine-tuning the model for a single aspect takes around 15 minutes.

The models trained with a multitask objective (\singletaskaae{}, \multitask{} and \multitaskaae{}) were fine-tuned for three epochs on a single A100-SXM4-80GB GPU, usiing a batch size of 64. With this setup, fine-tuning takes around 50 minutes for \singletaskaae{}, around 2 hours for \multitask{}, and around 3 hours for \multitaskaae{}.

We base our implementation of the multitask learning models on \url{https://github.com/shahrukhx01/multitask-learning-transformers}, as we found this to work notably better than alternative libraries.

For all models, we report results for a single training and inference run.

\subsection{Significance Tests}
Due to varying experimental settings, we employ different techniques to test for significance. Below, we describe and justify the applied testing methodology for each setting.

\paragraph{AAE Classification}
For the AAE classification presented in Table \ref{tab:results-aae-classification-subsampled}, we compare the results of the proposed classifiers to the TwitterAAE approach proposed by \citet{blodgett2016}. Since the code for the baseline is available, we are also able to retrieve per-sample predictions on the test dataset. Therefore, we calculate significance levels using a one-sided independent $t$-test, marked with \dag{} for $p<0.05$ and \ddag{} for $p<0.01$. Here, we employ a one-sided dependent paired $t$-test if the scores seem to be drawn from a normal distribution, and the Wilcoxon signed-rank test otherwise (as suggested in \citet{dror2018}, we test for normality using the Shapiro-Wilk test with $\alpha = 0.05$). To do so, we split test set of the TwitterAAE corpus (cf. Section \ref{sec:data}) into ten random subsets (for the \textit{Test smp} set described in Section~\ref{sec:results} and Table~\ref{tab:twitteraae-corpus-statistics}, this results in 45,991 instances per subset), calculate precision, recall and F$_1$-score for each subset and use the score distribution as input to the $t$-test.

\input{table-twitteraae-corpus-statistics}

\input{table-sbic-corpus-statistics}

\paragraph{Overall Social Bias Classification}
For the overall bias classification presented in Table \ref{tab:results-bias-classification}, we calculate two significance levels. First, we compare the results of the evaluated approaches to baselines from the literature, marked with \dag{} for $p<0.05$ and \ddag{} for $p<0.01$. As neither, the code nor per-sample predictions are available for either baseline at the time of writing, we employ a one-sample $t$-test. Since baseline scores are not available for the negative class and macro averaged F$_1$-scores, we cannot compute the significance over the baselines for results presented in Table \ref{tab:results-bias-classification-negative-class} and Table \ref{tab:results-bias-classification-macro}.

Second, we compare the multitask approaches to their respective single-task variants, marked with * for $p<0.05$ and ** for $p<0.01$. Since we train five models with five different random seeds, we calculate the F$_1$-score for each model seed and use the score distribution as input to the significance test.

\paragraph{Per Dialect Social Bias Classification}
For the classification results per dialect presented in Table \ref{tab:results-per-dialect-bias-classification}, we calculate two significance levels. First, we compare the multitask approaches to their respective single-task variants, marked with * for $p<0.05$ and ** for $p<0.01$. Second, we compare the results of the approaches that model AAE dialect to their respective non-AAE variants with \dag{} for $p<0.05$ and \ddag{} for $p<0.01$.

In both scenarios, we employ a one-sided dependent paired $t$-test. Since we train five models with five different random seeds, we calculate the F$_1$-score for each model seed and use the score distribution as input to the significance test.

%% file: table-twitteraae-corpus-statistics.tex
\begin{table}
    \small
    \centering
    \setlength{\tabcolsep}{3.2pt}

    \begin{tabular}{lrrrr}
        \toprule

        \bf Dialect & \bf Train & \bf Validation & \bf Test & \bf Test smp \\

        \midrule

        No-AAE & 37,171,287 & 9,292,822 & 11,616,028 & 229,955 \\
        AAE & 735,856 & 183,964 & 229,955 & 229,955 \\
        [0.5em]

        Total & 37,907,143 & 9,476,786 & 11,845,983 & 459,910 \\

        \bottomrule
    \end{tabular}

    \caption{The number of instances per split in the TwitterAAE dataset. The \emph{Test smp} column describes the numbers for the sampled test data used to evaluate the approaches, as presented in Section~\ref{sec:results}.}
    \label{tab:twitteraae-corpus-statistics}
\end{table}

%% file: table-sbic-corpus-statistics.tex
\begin{table*}
    \small
    \centering
    \setlength{\tabcolsep}{5pt}

    \begin{tabular}{lrrrrrrrrrrrr}
        \toprule

        & \multicolumn{4}{c}{\textbf{Train}} & \multicolumn{4}{c}{\textbf{Validation}} & \multicolumn{4}{c}{\textbf{Test}} \\
        \cmidrule(l@{2pt}r@{2pt}){2-5}\cmidrule(l@{2pt}r@{2pt}){6-9}\cmidrule(l@{2pt}r@{2pt}){10-13}

        & \multicolumn{2}{c}{\textbf{Positive}} & \multicolumn{2}{c}{\textbf{Negative}} & \multicolumn{2}{c}{\textbf{Positive}} & \multicolumn{2}{c}{\textbf{Negative}} & \multicolumn{2}{c}{\textbf{Positive}} & \multicolumn{2}{c}{\textbf{Negative}}  \\
        \cmidrule(l@{2pt}r@{2pt}){2-3}\cmidrule(l@{2pt}r@{2pt}){4-5}\cmidrule(l@{2pt}r@{2pt}){6-7}\cmidrule(l@{2pt}r@{2pt}){8-9}\cmidrule(l@{2pt}r@{2pt}){10-11}\cmidrule(l@{2pt}r@{2pt}){12-13}

        \textbf{Label} & \textbf{$\lnot$AAE} & \textbf{AAE} & \textbf{$\lnot$AAE} & \textbf{AAE} & \textbf{$\lnot$AAE} & \textbf{AAE} & \textbf{$\lnot$AAE} & \textbf{AAE} & \textbf{$\lnot$AAE} & \textbf{AAE} & \textbf{$\lnot$AAE} & \textbf{AAE} \\

        \midrule

        Offensiveness & 16294 & 2432 & 15286 & 1492 & 2281 & 331 & 1874 & 187 & 2342 & 368 & 1797 & 191 \\
        Intent & 14795 & 2181 & 16785 & 1743 & 2109 & 306 & 2046 & 212 & 2171 & 326 & 1968 & 233 \\
        Lewdness & 3092 & 497 & 28488 & 3427 & 365 & 56 & 3790 & 462 & 383 & 69 & 3756 & 490 \\
        Target Group & 10624 & 1556 & 20956 & 2368 & 1581 & 234 & 2574 & 284 & 1690 & 245 & 2449 & 314 \\
        Ingroup & 647 & 321 & 30933 & 3603 & 56 & 41 & 4099 & 477 & 4088 & 38 & 51 & 521 \\

        \bottomrule
    \end{tabular}

    \caption{The number of instances per split, label, and dialect in the Social Bias Inference Corpus \cite{sap2020}. The dialect labels were inferred automatically using the approach presented in Section~\ref{sec:method}.}
    \label{tab:sbic-corpus-statistics}
\end{table*}

%% file: acl24-dialect-bias-appendix-d.tex
\section{Dataset Details}
\label{sec:appendix-data-details}

\subsection{TwitterAAE}
Detailed dataset statistics of the TwitterAAE Corpus \cite{blodgett2016} are reported in Table~\ref{tab:twitteraae-corpus-statistics}.

\subsection{Social Bias Inference Corpus}
Detailed dataset statistics of the Social Bias Inference Corpus \cite{sap2020} are reported in Table~\ref{tab:sbic-corpus-statistics}. Dialect labeles are inferred automatically using the \aaeweightedloss{} approach presented in Section~\ref{sec:method}.